\documentclass{article}

\usepackage{microtype}
\usepackage{graphicx}
\usepackage{booktabs}
\usepackage{hyperref}

\usepackage[accepted]{icml2026}

\usepackage{newtxtext}

\usepackage{amsmath}
\usepackage{amssymb}
\usepackage{enumitem}
\usepackage{placeins}

\icmltitlerunning{Structured Hints for Lean Provers}

\begin{document}

\twocolumn[
\icmltitle{Inference-Time Diversity in\\RL-Trained Lean Theorem Provers:\\A Diagnostic Study}

\begin{icmlauthorlist}
\icmlauthor{Zachary Burton}{mit}
\end{icmlauthorlist}

\icmlaffiliation{mit}{Massachusetts Institute of Technology, Cambridge, MA, USA}

\icmlcorrespondingauthor{Zachary Burton}{zacburton@alum.mit.edu}

\icmlkeywords{Neural Theorem Proving, Lean4, Reinforcement Learning}

\vskip 0.3in
]

\printAffiliationsAndNotice{}

% icml2026.sty's height threshold (6.25pt) is below the ascender height of
% \small\bf, so the running-head check fires for any non-empty title. Force the
% header here to bypass the false positive.
\makeatletter
\gdef\@icmltitlerunning{Inference-Time Diversity in Lean Provers}
\makeatother
\fancyhead[C]{\small\bf Inference-Time Diversity in Lean Provers}

\begin{abstract}
RL-trained Lean theorem provers mode-collapse at inference time: on miniF2F-test with DeepSeek-Prover-V1.5-RL, doubling the i.i.d.\ sampling budget from $k{=}32$ to $k{=}64$ produces zero additional solved theorems (42/244 in both cases). A fixed schedule of 15 tactic skeletons breaks this plateau and recovers a $+45\%$ relative improvement at $k{=}16$ (mean $\Delta = +12.3 \pm 4.2$ theorems across $n{=}3$ seeds, sign preserved in every seed). A controlled diversity ablation rules out the prompt-diversity confound: tactic skeletons help, paraphrases match the baseline, and irrelevant Lean comments actively degrade. A leave-one-out formalization-difficulty stratification reveals a structural-content gradient across the three perturbations. The phenomenon is RL-specific: V1.5-Base proves zero theorems regardless of intervention, identifying RL as the stage that creates the proof capability which subsequently collapses; extending to two additional 7B Lean provers, RL-trained DeepSeek-Prover-V2-7B contributes $+3$ frontier solves no i.i.d.\ baseline can reach despite a flat aggregate, while SFT-trained Goedel-Prover does not ($-10.0 \pm 4.4$ theorems, $n{=}3$, sign preserved every seed). Inference-time structural diversity is a cheap, complementary axis for RL-trained provers, orthogonal to scaling model size or training compute.
\end{abstract}

\section{Introduction}

Recent neural theorem provers combine large language models with reinforcement learning and tree searches to achieve strong results on formal proof benchmarks~\cite{xin2024deepseek, chen2025seed, trinh2024solving, lin2025goedel}.\footnote{Code, prompts, and per-attempt logs: \url{https://github.com/zacn04/icml_hints_for_provers}.} The prevailing paradigm assumes that sufficient RL training and inference-time compute will cause models to internalize the structural priors required for formal proofs.

We present evidence that this assumption is flawed. RL-trained provers suffer from \emph{mode collapse} at inference time: they converge on a narrow set of proof strategies and fail to explore alternatives even when given additional sampling budget. On miniF2F-test with DeepSeek-Prover-V1.5-RL, i.i.d.\ sampling at temperature 0.6 solves 38 theorems at $k{=}16$, 42 at $k{=}32$, and \emph{still 42} at $k{=}64$---32 additional stochastic samples find zero new proofs.

A fixed prompt schedule of 15 common tactic skeletons (\texttt{simp}, \texttt{intro}, \texttt{constructor}, \dots) breaks the plateau, yielding a $+45\%$ relative improvement at $k{=}16$ (55 vs.\ 38) and continuing to find new proofs through $k{=}64$. A controlled diversity ablation (skeleton vs.\ paraphrase vs.\ irrelevant comment) confirms the gains are \emph{structural}, not an artifact of prompt diversity. Stratifying by an empirical formalization-difficulty score grounded in baseline solvability across our experimental matrix further localizes the effect.

A recent line of work has examined whether RL with verifiable rewards (RLVR) genuinely expands a base model's reasoning capacity~\cite{yue2025rlvr, wu2025invisibleleash}. We study a different question---inference-time collapse within an RL-trained policy in an exact-verifier domain---and discuss the relationship in detail in Section~\ref{sec:related}.

\subsection{Contributions}
\begin{itemize}
    \item We diagnose mode collapse in RL-trained Lean provers via two complementary signatures: an i.i.d.\ sampling plateau at $k{=}32\to64$ on V1.5-RL that a fixed 15-skeleton schedule breaks, and a controlled diversity ablation (skeleton vs.\ paraphrase vs.\ irrelevant comment) showing the gain comes from structural guidance, not prompt diversity. A dedicated NL-only ablation (C3) further isolates the natural-language hint contribution from the tactic-prefix contribution.
    \item A leave-one-out empirical-difficulty stratification, grounded in cross-prover baseline solvability, localizes the structural gain to the easy/trivial buckets where V1.5-RL's gap to other open-source provers concentrates, and reveals a clean structural-content gradient across the three perturbations.
    \item We establish RL-specificity in two ways. A within-architecture base-vs-RL contrast on V1.5 shows the non-RL base has \emph{zero} proof capability, so RL is both the source of the capability and the source of its collapse. A cross-model contrast on two additional 7B provers shows that RL-trained DeepSeek-Prover-V2-7B gains $+3$ frontier theorems unreachable by any i.i.d.\ baseline despite a flat aggregate, while SFT-trained Goedel-Prover loses $-10.0 \pm 4.4$ theorems across $n{=}3$ seeds (sign preserved every seed).
\end{itemize}

\section{Related Work}
\label{sec:related}

\paragraph{Neural theorem provers.}
The integration of language models with formal proof assistants has progressed quickly. Early systems like GPT-f~\cite{polu2020generative} and PACT~\cite{han2022proof} demonstrated that transformers could generate tactic scripts, and LeanDojo with ReProver~\cite{yang2023leandojo} added retrieval to handle Lean's large premise library. More recent provers combine reinforcement learning with tree search to push miniF2F performance above 60\%: the DeepSeek-Prover family (V1.5~\cite{xin2024deepseek} and V2~\cite{deepseekv2prover}) and the supervised-fine-tuned Goedel-Prover~\cite{lin2025goedel}. We deliberately anchor our analysis on V1.5-RL: it is the only widely benchmarked RL-trained Lean prover that also releases a non-RL base variant, which is what makes the clean within-architecture isolation of RL's effect possible in Section~\ref{sec:base_vs_rl}.

\paragraph{Tree search and hint-based methods.}
HyperTree Proof Search~\cite{lample2022hypertree} and COPRA~\cite{thakur2023copra} build proof trees one tactic at a time with explicit backtracking; our skeleton schedule can be viewed as fixing only the \emph{first} node of the proof tree and letting the model complete the rest in a single forward pass with no proof-state feedback. Draft-Sketch-Prove~\cite{jiang2023dsp}, ConjectureBench~\cite{sivakumar2025conjecturing}, and Aristotle~\cite{achim2025aristotle} intervene at a higher level of abstraction (informal sketches, conjecture generation, intermediate semantic lemmas), whereas we operate at the syntactic tactic-stem level and focus on \emph{diagnosing} RL-induced underexploration rather than proposing a new proving architecture. Lean~4~\cite{demoura2021lean4} and Mathlib~\cite{mathlib} form our verification stack and miniF2F~\cite{zheng2022minif2f} is our benchmark throughout.

\paragraph{Entropy collapse and the limits of RL post-training.}
RL post-training narrows the output distribution of the underlying base model, trading diversity against best-of-$N$ performance~\cite{kirk2024understanding}. Two recent works study whether RL with verifiable rewards (RLVR) expands a base model's reasoning support: \citet{yue2025rlvr} find that base models surpass their RL-trained variants at large $k$ on math and coding benchmarks with answer-string verification, and \citet{wu2025invisibleleash} argue more generally that the contraction of output support induced by RLVR can outweigh its expansion of high-reward modes. Our setting differs from those measurements in two relevant ways: the Lean kernel verifies proofs exactly rather than by string-match or unit-test (so V1.5-Base proves zero theorems at any $k$, Section~\ref{sec:base_vs_rl}), and our cross-model comparison includes an SFT-trained control (Goedel-Prover) absent from those setups.

\section{Method}

\subsection{Skeleton Schedule and Ablations}
\label{sec:methods_schedule}

A structured query is the pair $(x, s)$ of theorem statement $x$ and tactic skeleton $s$. The schedule is a deterministic $15{\times}8{=}120$ grid: 15 hand-selected Lean tactic skeletons (Appendix~\ref{app:skeletons}, including the empty skeleton) crossed with 8 goal-hint comments (the first being empty). Schedule advances tactic-first, hint-second; each of the 120 attempts is a distinct (skeleton, hint) prompt configuration with no repeats within $k{\leq}120$. Our tested budgets $k\in\{16,32,64\}$ sample the first 16, 32, and 64 grid entries. The schedule is fully deterministic; stochasticity comes only from temperature sampling.

To isolate structural content from prompt diversity per se, we add two ablations (Appendix~\ref{app:ablation}): \textbf{C1 (paraphrase)}---16 semantically equivalent rephrasings of the instruction injected as Lean comments, with empty tactic prefix; and \textbf{C2 (comment)}---16 content-free Lean comments (\texttt{/- approach alpha -/}, \dots) prepended to the theorem.

\subsection{Experimental Setup}
\label{sec:setup}

\paragraph{Models.}
The primary within-architecture analysis (Sections~\ref{sec:scaling}--\ref{sec:base_vs_rl}) uses DeepSeek-Prover-V1.5-RL together with its non-RL base variant V1.5-Base---same pretraining corpus, no RL fine-tuning stage. V1.5 is the only widely benchmarked RL-trained Lean prover that releases a paired non-RL checkpoint, which is what enables the clean within-architecture isolation of RL's effect. The cross-model study in Section~\ref{sec:multi_model} extends the pass@16 comparison to two additional 7B provers: DeepSeek-Prover-V2~\cite{deepseekv2prover}, RL-trained and emitting chain-of-thought before the Lean output; and Goedel-Prover-SFT~\cite{lin2025goedel}, which is supervised-fine-tuned with no RL stage and emits Lean code directly. Neither additional model releases a paired non-RL or non-SFT base checkpoint, so the within-architecture RL-vs-no-RL contrast remains possible only on V1.5.

\paragraph{Skeleton delivery.}
We deliver the structural intervention in each model's native interaction modality. For completion-mode provers (V1.5-RL, V1.5-Base, Goedel-Prover) this is a literal Lean tactic prefix injected after \texttt{:= by}. For the reasoning-mode prover (V2) it is a natural-language instruction in the chat prompt: \emph{``Begin the proof with: \texttt{<tactic>}.''} Both delivery mechanisms operationalize the same underlying hypothesis---that the model should commence the proof from a specific tactic stem---adapted to the prompting paradigm each model was trained for. Forcing a literal Lean prefix on a chat-mode reasoning model would conflict with its native chain-of-thought process and conflate the structural-guidance claim with a separate question about cross-paradigm prompt-format robustness.

\paragraph{Search procedure.}
Both conditions execute exactly $k$ LLM calls per theorem and the search exits on the first attempt that produces a Lean-verified proof. Pass@$k$ is reported in the strict sense: any of the $k$ attempts succeeds. Sample mode ($A$) holds the prompt fixed (empty) and varies the random seed across attempts; structured mode ($B$) holds the seed fixed and varies the prompt across attempts according to the schedule. Both modes consume an identical LLM-call budget at $k{=}16$, making the comparison an apples-to-apples test of two sources of diversity. Model outputs are post-processed before Lean verification (theorem-signature stripping, markdown-fence removal, re-indentation); see Appendix~\ref{app:env} for the exact pipeline.

\paragraph{Benchmark and environment.}
We evaluate on miniF2F-test (244 theorems, Lean 4 split), with the Lean toolchain pinned to v4.9.0-rc1 and Mathlib pinned to commit \texttt{7fa489a5}. This pinning matches the environment used to train DeepSeek-Prover-V1.5 and is critical for clean attribution: newer Lean versions ship substantially more powerful built-in tactic solvers (notably \texttt{simp}, \texttt{aesop}, and \texttt{nlinarith}), and running on those toolchains would conflate the gain from our intervention with the gain from Lean's own evolution.

\paragraph{Decoding and inference.}
Sampling uses temperature 0.6 and top-$p$ 0.95 for all conditions. Completion-mode provers use a 1024-token generation budget per attempt; reasoning-mode provers use 8192 tokens to accommodate their chain-of-thought before the Lean code block. Inference runs on a single A100 GPU per experiment (80GB for the V1.5 runs, 40GB for V2 and Goedel-Prover; both sufficient). Lean verification uses \texttt{lake env lean --json} with a 120-second per-attempt timeout, and we shard theorems 8-way across parallel workers ($i \bmod 8$). Full code, prompt templates, and per-attempt logs are available at \url{https://github.com/zacn04/icml_hints_for_provers}.

\section{Results}

\subsection{Scaling Analysis: Mode Collapse in i.i.d.\ Sampling}
\label{sec:scaling}

Table~\ref{tab:scaling} presents our main finding. The baseline (i.i.d.\ sampling) plateaus at $k{=}32$: doubling the budget to $k{=}64$ produces zero additional solved theorems. In contrast, the skeleton-guided approach continues to find new proofs at each budget level.

\begin{table*}[t]
\centering
\begin{tabular}{lccc}
\toprule
\textbf{Condition} & \textbf{Pass@16} & \textbf{Pass@32} & \textbf{Pass@64} \\
\midrule
A-RL (i.i.d.\ baseline) & 38/244 (15.6\%) & 42/244 (17.2\%) & 42/244 (17.2\%) \\
\textbf{B-RL (skeleton schedule)} & \textbf{55/244 (22.5\%)} & \textbf{58/244 (23.8\%)} & \textbf{60/244 (24.6\%)} \\
\midrule
Absolute gap & +17 & +16 & +18 \\
Relative improvement & +44.7\% & +38.1\% & +42.9\% \\
\bottomrule
\end{tabular}
\caption{Scaling analysis on miniF2F-test with DeepSeek-Prover-V1.5-RL. The i.i.d.\ baseline plateaus at $k{=}32$ while the skeleton schedule continues to find new proofs. The gap is persistent across all budget levels.}
\label{tab:scaling}
\end{table*}

\begin{figure}[t]
\centering
\includegraphics[width=\columnwidth]{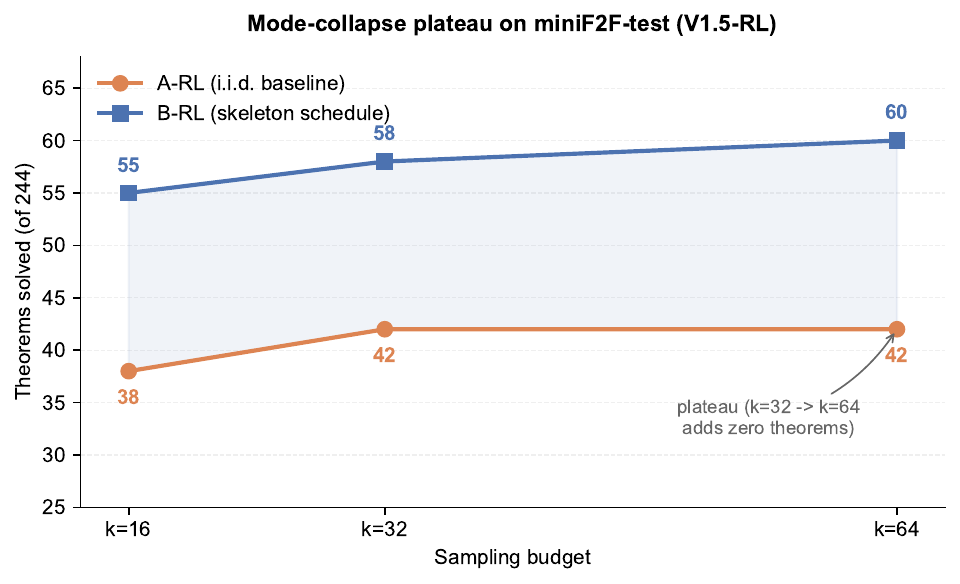}
\caption{A-RL i.i.d.\ sampling flatlines at 42/244 from $k{=}32$ onward; B-RL with the 15-skeleton schedule keeps climbing.}
\label{fig:scaling}
\end{figure}

The baseline's plateau from $k{=}32$ to $k{=}64$ is striking: 32 additional samples---each with different random seeds---fail to find a single new proof. This indicates that the model is trapped in a narrow region of the proof strategy space. Temperature sampling alone cannot break this mode collapse.

\subsection{Diversity Ablation: Structure vs.\ Diversity}
\label{sec:ablation}

Table~\ref{tab:ablation} compares four conditions at $k{=}16$ to isolate the source of improvement.

\begin{table}[t]
\centering
\small
\begin{tabular}{lcc}
\toprule
\textbf{Condition} & \textbf{Solved} & \textbf{Pass@16} \\
\midrule
C2 (irrelevant comments)   & 25 & 10.2\% \\
A-RL (i.i.d.\ baseline)    & 38 & 15.6\% \\
C1 (instruction paraphrases) & 38 & 15.6\% \\
\textbf{B-RL (skeletons)}  & \textbf{55} & \textbf{22.5\%} \\
\bottomrule
\end{tabular}
\caption{Diversity ablation at $k{=}16$ (all $/244$). Skeletons improve; paraphrases match baseline at topline; comments degrade. Section~\ref{sec:difficulty} shows C1's match hides redistribution.}
\label{tab:ablation}
\end{table}

\begin{figure}[t]
\centering
\includegraphics[width=\columnwidth]{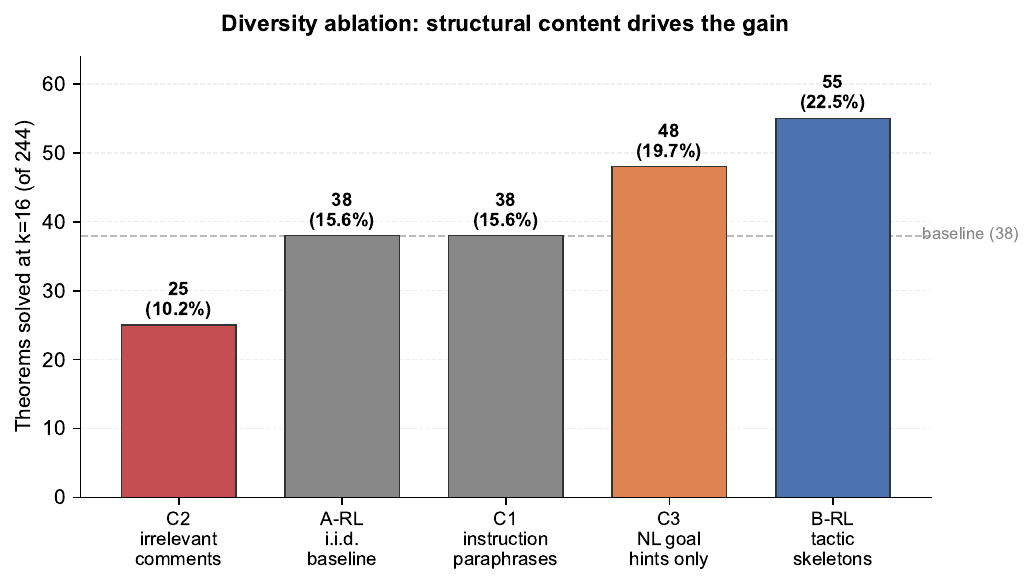}
\caption{Structural-content gradient at $k{=}16$: C2 $<$ A-RL $=$ C1 $<$ C3 $<$ B-RL. Irrelevant tokens hurt; pure paraphrase matches the baseline; NL goal hints alone (C3) recover most of the A--B gap; tactic stems go furthest.}
\label{fig:ablation}
\end{figure}

The C2 result is interesting. We observe that irrelevant Lean comments injected before the theorem actually \emph{degrade} performance, suggesting that surface-level prompt perturbations that lack tactic-level structure interfere with the model's collapsed policy. The C1 topline equality with A-RL is suggestive but, by itself, ambiguous: it is consistent with both ``C1 has no effect'' and ``C1 helps on some theorems while hurting an equal number on others.'' We disambiguate these in Section~\ref{sec:difficulty} by stratifying solves by empirical formalization difficulty.

\subsection{Empirical Difficulty Stratification}
\label{sec:difficulty}

Mathematical difficulty (e.g., IMO vs.\ textbook) is a poor proxy for the difficulty a neural prover actually faces in Lean, because the choice of formalization can \textbf{materially shift the proof difficulty independently of the underlying mathematics}. We therefore define an \emph{empirical formalization-difficulty} score per theorem, grounded in what current ML provers can actually do on these specific Lean formalizations.

\paragraph{Difficulty score (leave-one-out).}
For each theorem $\tau$ and a target model $M$ under analysis, we define $\sigma_{-M}(\tau)$ as the number of hint-free i.i.d.\ sampling baselines from \emph{other} models that solve $\tau$. This leave-one-out construction prevents the score from being contaminated by $M$'s own sample runs, which also serve as the reference for the ``NEW'' analysis. For analyses of V1.5-RL, the four baselines used to compute $\sigma_{-\text{V1.5-RL}}$ are Goedel-Prover sample at $k{\in}\{16,32\}$ and V2-7B sample at $k{\in}\{16,32\}$. (An equivalent leave-one-out score is used for the V2 analysis in Section~\ref{sec:multi_model}.) Theorems with $\sigma_{-M}{=}0$ are beyond the reach of every \emph{other} baseline at every tested $k$; theorems with $\sigma_{-M}{=}N_{-M}$ (the maximum, 4 for V1.5-RL) are solved by every other baseline. The distribution under $\sigma_{-\text{V1.5-RL}}$ is reported in Table~\ref{tab:difficulty_dist}.

\begin{table}[t]
\centering
\small
\begin{tabular}{lc}
\toprule
\textbf{Bucket} & \textbf{\#} \\
\midrule
Frontier ($\sigma{=}0$)  & 117 \\
Hard     ($\sigma{=}1$)  & 24  \\
Medium   ($\sigma{=}2$)  & 13  \\
Easy     ($\sigma{=}3$)  & 53  \\
Trivial  ($\sigma{=}4$)  & 37  \\
\midrule
\textbf{Total}           & \textbf{244} \\
\bottomrule
\end{tabular}
\caption{Leave-one-out difficulty distribution of miniF2F-test for V1.5-RL analyses. $\sigma = \sigma_{-\text{V1.5-RL}}$ counts how many of four non-V1.5-RL sample baselines (Goedel $\times 2$, V2 $\times 2$) solve the theorem. 48\% are in the frontier bucket.}
\label{tab:difficulty_dist}
\end{table}

\paragraph{Where the perturbations operate.}
For each condition $C$, let $\text{NEW}(C)$ denote the set of theorems solved by $C$ but not by V1.5-RL sample at $k{=}16$, and let $\text{LOST}(C)$ denote the set solved by sample but not by $C$. Stratifying these sets by $\sigma_{-\text{V1.5-RL}}$ yields Table~\ref{tab:bucket_strat}.

\begin{table*}[t]
\centering
\small
\begin{tabular}{lcccccc}
\toprule
\textbf{Condition} & \textbf{Frontier} & \textbf{Hard} & \textbf{Medium} & \textbf{Easy} & \textbf{Trivial} & \textbf{NET} \\
\midrule
Skeleton (B-RL) & 0 & $+1$ & $+1$  & $\mathbf{+9}$  & $\mathbf{+6}$  & $\mathbf{+17}$ \\
Paraphrase (C1) & 0 & 0    & 0     & $+3$           & $-3$           & 0              \\
Comment (C2)    & 0 & $+1$ & 0     & $-6$           & $-8$           & $-13$          \\
\bottomrule
\end{tabular}
\caption{NET effect ($\text{NEW}-\text{LOST}$) of each V1.5-RL perturbation at $k{=}16$, stratified by $\sigma_{-\text{V1.5-RL}}$. The skeleton intervention's $+17$ net is dominated by recoveries in the \emph{easy} ($+9$) and \emph{trivial} ($+6$) buckets---theorems that other open-source provers routinely solve via sample but that V1.5-RL sample misses, exactly the population we would expect a mode-collapsed model to fail on. Paraphrase (C1) shows only a small redistribution between easy and trivial; comment (C2) shows uniform degradation across easy and trivial. \textbf{No perturbation moves any theorem out of the $\sigma_{-\text{V1.5-RL}}{=}0$ frontier.}}
\label{tab:bucket_strat}
\end{table*}

\paragraph{What the stratification reveals.}
The skeleton intervention specifically remediates V1.5-RL's gap to the rest of the field. Of its 17 net wins, 15 fall in the easy ($+9$) and trivial ($+6$) buckets---that is, on theorems where three or four of the four other-model baselines already solve without any intervention. This is the operational signature of mode collapse: V1.5-RL is failing on problems that other RL- and SFT-trained provers solve routinely, and the skeleton intervention recovers precisely those problems.

The two non-structural perturbations behave very differently. Paraphrase (C1) reshuffles wins within the easy and trivial buckets without producing any net gain, and comment (C2) degrades performance uniformly across those same buckets. The structural-content gradient is sharpest in the easy-plus-trivial column, where the three perturbations sit at $+15$ (skeleton), $0$ (paraphrase), and $-14$ (comment).

Higher sample budgets do not change the frontier picture. Replicating the stratification at $k{=}32$ and $k{=}64$ on V1.5-RL, no perturbation unlocks a single $\sigma_{-\text{V1.5-RL}}{=}0$ frontier theorem at any tested budget. Some fraction of the 117 frontier theorems likely reflect known formalization issues in miniF2F-test---missing hypotheses or ambiguous statements on specific competition problems---rather than pure capability limits, but we cannot disentangle the two with the data we have.

\paragraph{Decomposing the skeleton win.}
The $k{=}16$ schedule includes one wrap attempt (attempt 15) where the tactic prefix is empty but a natural-language goal-hint comment (e.g., \texttt{/-- Hint: Start by simplifying using simp. --/}) is prepended. We count V1.5-RL structured wins by attempt class (Table~\ref{tab:attempt_class}).

\begin{table}[t]
\centering
\small
\begin{tabular}{lcc}
\toprule
\textbf{Attempt class} & \textbf{Wins} & \textbf{NEW} \\
\midrule
Attempt 0 (truly empty)           & 16 & 0  \\
Attempt 15 (hint comment only)    & 9  & 8  \\
Tactic prefix (attempts 1--14)    & 30 & 12 \\
\midrule
\textbf{Total}                    & \textbf{55} & \textbf{20} \\
\bottomrule
\end{tabular}
\caption{V1.5-RL structured $k{=}16$ wins by attempt class. ``NEW'' = wins not in V1.5-RL sample $k{=}16$. The hint-comment slot contributes $+8$ NEW, tactic prefixes contribute $+12$. Dominant tactic prefixes: \texttt{intros} ($+6$), \texttt{norm\_num} ($+3$).}
\label{tab:attempt_class}
\end{table}

The 0/16 NEW count for attempt-0 confirms V1.5-RL is effectively deterministic at our inference settings---the $+17$ topline cannot be attributed to vLLM sampling variance. The structural intervention works through two delivery layers: natural-language hint comment ($+8$ NEW) and literal tactic prefix ($+12$ NEW), both contributing additively to the $+20$ total.

\paragraph{Dedicated NL-only ablation (C3).}
The attempt-class decomposition pins the NL hint contribution to a single slot of the composite schedule, leaving open whether the effect generalizes when the entire budget is spent on NL comments. We therefore ran a dedicated condition C3 in which all 16 attempts on V1.5-RL use empty tactic prefixes and cycle through the 7 distinct natural-language goal-hint comments (Appendix~\ref{app:skeletons}), with no tactic stem injected at any attempt. C3 solves $48/244\,(19.7\%)$---intermediate between sample-only A ($15.6\%$) and structured-skeleton B ($22.5\%$). The clean ordering $A < C3 < B$ confirms the structural-content gradient hypothesized in Section~\ref{sec:setup}: NL hints alone recover roughly $58\%$ of the A$\to$B gap ($+10$ solves of the $+17$ B-A topline), and tactic stems contribute the orthogonal remainder ($+7$). Skeleton guidance is therefore not reducible to either modality alone---both layers contribute, and the two effects compose.

\subsection{Base vs.\ RL: The Origin of Mode Collapse}
\label{sec:base_vs_rl}

\begin{table*}[t]
\centering
\begin{tabular}{lccc}
\toprule
\textbf{Condition} & \textbf{Pass@16} & \textbf{Pass@32} & \textbf{Pass@64} \\
\midrule
A-BASE (i.i.d.) & 0/244 (0.0\%) & 0/244 (0.0\%) & 0/244 (0.0\%) \\
B-BASE (skeletons) & 0/244 (0.0\%) & 0/244 (0.0\%) & 0/244 (0.0\%) \\
\midrule
A-RL (i.i.d.) & 38/244 (15.6\%) & 42/244 (17.2\%) & 42/244 (17.2\%) \\
B-RL (skeletons) & 55/244 (22.5\%) & 58/244 (23.8\%) & 60/244 (24.6\%) \\
\bottomrule
\end{tabular}
\caption{Base model vs.\ RL model on miniF2F-test. The base model proves zero theorems across all conditions and budgets---inspection of generated outputs shows it almost always produces \texttt{sorry} or echoes the theorem statement back, indicating it has not learned to generate valid Lean 4 proofs at all. Mode collapse is therefore an RL-specific phenomenon: RL is what creates proof capability in the first place, and our intervention recovers proofs that the RL-collapsed model fails to find on its own.}
\label{tab:base}
\end{table*}

The base model result is unambiguous and somewhat surprising: \textbf{DeepSeek-Prover-V1.5-Base proves zero theorems on miniF2F-test}, regardless of sampling budget or whether tactic skeletons are provided. Inspection of the generated outputs across all 244 theorems and all 64 attempts reveals that the base model produces \texttt{sorry} in 58.8\% of its outputs, or echoes the theorem statement verbatim without attempting a proof. By contrast, the RL model almost never generates explicit \texttt{sorry} tokens ($<$0.2\% under i.i.d.\ sampling); its mode collapse manifests instead as persistent generation of failing but syntactically valid tactic sequences---repeated \texttt{rw} or \texttt{apply} attempts that do not close the goal.

This reframes our hypothesis. We had originally predicted that the base model would show a \emph{smaller} skeleton-induced improvement, providing graded evidence that RL exacerbates collapse. Instead, the base model shows no improvement because it has \emph{no proof capability to collapse}: the RL fine-tuning stage is what teaches the model to generate valid Lean 4 tactic proofs in the first place. Skeletons cannot substitute for this capability, nor goad out theorem proving knowledge where it is non-existent.

The implication is sharper than an earlier framing we posited: mode collapse is not merely \emph{exacerbated} by RL training, it is \emph{caused by} it. RL creates a useful but narrow proof policy; without the structural intervention provided by tactic skeletons, the model cannot escape the policy learned at training time, even when it may benefit. The base model demonstrates that the underlying language model has no useful tactic-level prior: the RL stage is the locus of both proof capability and the inference-time collapse that limits it.

\subsection{Generalization Across Models}
\label{sec:multi_model}

The V1.5 RL-vs-Base contrast in Section~\ref{sec:base_vs_rl} pins mode collapse to the RL stage \emph{within} one model family. To test whether the same A-vs-B pattern generalizes across architectures and training pipelines, we extend the pass@16 comparison to two additional 7B Lean provers: a further RL-trained model (DeepSeek-Prover-V2-7B) and one SFT-only model (Goedel-Prover).

\begin{table*}[t]
\centering
\small
\begin{tabular}{llccc}
\toprule
\textbf{Model} & \textbf{Post-Training} & \textbf{A pass@16} & \textbf{B pass@16} & \textbf{$\Delta$} \\
\midrule
V1.5-Base    & none (pre-train only) & 0/244 (0.0\%)    & 0/244 (0.0\%)    & 0    \\
Goedel-Prover & SFT                  & 90/244 (36.9\%)  & 85/244 (34.8\%)  & $-5$ \\
\midrule
V1.5-RL      & RL                    & 38/244 (15.6\%)  & \textbf{55/244 (22.5\%)} & $+17$ \\
V2-7B        & RL                    & 126/244 (51.6\%) & \textbf{127/244 (52.0\%)} & $+1$ \\
\bottomrule
\end{tabular}
\caption{Cross-model A vs.\ B comparison at $k{=}16$ on miniF2F-test. The aggregate $\Delta$ varies by model: V1.5-RL shows the largest lift (+17), V2-7B is flat on aggregate (+1) but solves $+3$ frontier theorems that no i.i.d.\ baseline reaches (Section~\ref{sec:multi_model}), and SFT-trained Goedel-Prover shows $-5$. Together with V1.5-Base (which has no proof capability with or without hints), the pattern is consistent with the hypothesis that structured hints remedy the underexploration induced by RL fine-tuning; the magnitude of remediation varies with the post-training recipe.}
\label{tab:cross_model}
\end{table*}

\paragraph{The Goedel-Prover null is diagnostic, not contradictory.} Goedel-Prover-SFT achieves the highest absolute pass@16 of any open-source prover in our study (36.9\% seed=1; mean $97.7 \pm 6.7$ theorems across $n{=}3$ seeds, Section~\ref{sec:seed}) using purely supervised fine-tuning on a large synthetic Lean corpus, with no RL stage. Despite this capability advantage over V1.5-RL, structured hints do not improve its performance---they cost it $-10.0 \pm 4.4$ theorems on average across the three seeds, with the sign preserved in every seed. Goedel also shows the same k-scaling plateau we observed for V1.5-RL: pass@16 and pass@32 are identical at 36.9\% for the i.i.d.\ baseline, and the structured condition scales similarly (34.8\% $\to$ 36.1\%). The \emph{plateau itself} thus appears to be a general property of i.i.d.\ sampling from these models, but the \emph{remediability of the plateau via structured hints} is specific to RL-trained models.

\paragraph{A plausible interpretation.} SFT and RL induce different modes of underexploration. SFT pattern-matches a model onto a fixed distribution of training proofs; mode collapse for such a model manifests as tactic-local exploration that already starts from a broad set of stems---skeletons add nothing because the model would have tried those stems anyway. RL reward-shapes a model onto a narrower set of high-reward strategies and abandons alternative tactic stems entirely; skeletons help because they force commencement from stems the RL policy has been trained away from. The asymmetry we observe is consistent with this account, but the magnitude varies with the RL training pipeline. V1.5-RL shows the cleanest aggregate gain ($+17$); V2-7B is flat on the aggregate but contributes $+3$ frontier solves that no i.i.d.\ baseline of any model in our matrix reaches---a structurally meaningful gain that the topline metric hides. Goedel-Prover (SFT-trained) shows a small negative aggregate. We read these together as evidence that RL-trained provers are reachable through structural intervention in a way SFT-only provers are not, while acknowledging that the size of the gain is contingent on the specifics of the post-training recipe and likely on how much capability the i.i.d.\ baseline has already saturated.

\paragraph{V2 frontier solves.} The $+1$ aggregate hides the most theoretically interesting outcome in the cross-model matrix. We compute $\sigma_{-\text{V2}}(\tau)$ on the same leave-one-out construction as Section~\ref{sec:setup}, dropping all V2 baselines from the difficulty score. Three theorems with $\sigma_{-\text{V2}}{=}0$---\texttt{imo\_1960\_p2}, \texttt{imo\_1962\_p2}, and \texttt{numbertheory\_x5neqy2p4}---are solved by V2 structured at $k{=}16$ but by none of the other six sample baselines in our matrix (V1.5-RL, Goedel, V2-sample, at multiple $k$). Two are IMO problems. These are theorems where structured guidance is the only known route to a solution at this compute scale, and they are exactly the kind of frontier finding that the aggregate metric obscures and that V1.5-RL's structured condition does \emph{not} produce (Section~\ref{sec:difficulty}). V2 structured is therefore the first condition in our matrix that opens up the $\sigma{=}0$ frontier, even though its aggregate is flat. We refrain from inferring a general pattern from $n{=}3$, but flag this as evidence that the aggregate $\Delta$ metric is a lossy summary of where structural guidance actually pays off, and as a concrete falsifier for the strong version of our hypothesis: if frontier theorems were beyond reach of all current models in this size class, V2 structured should not have unlocked any.

\begin{figure}[t]
\centering
\includegraphics[width=0.85\columnwidth]{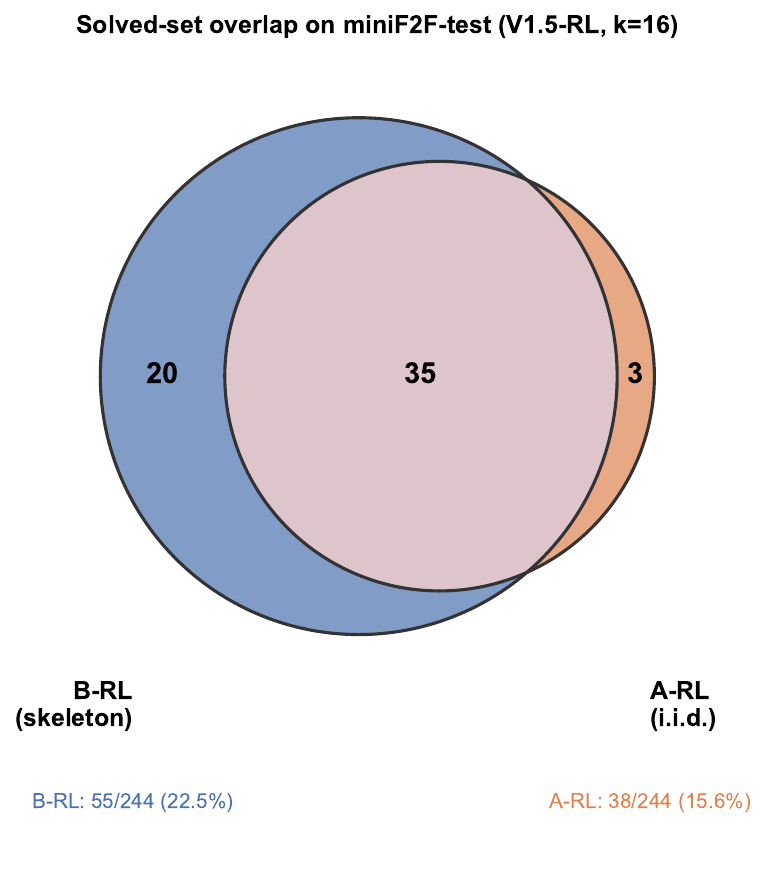}
\caption{Solved-set overlap of A-RL vs.\ B-RL at $k{=}16$ on V1.5-RL. 20 theorems are solved only by B-RL, 3 only by A-RL, 35 by both.}
\label{fig:venn}
\end{figure}

\paragraph{Solved-set overlap.} A paired view of A-RL vs.\ B-RL at $k{=}16$ further constrains the ``more compute would suffice'' objection (Figure~\ref{fig:venn}): of the 38 theorems solved by A-RL, B-RL solves 35 of them and adds 20 new ones (only 3 A-RL solves are lost, consistent with stochastic variation in shard scheduling). The intervention is therefore near-Pareto-dominant on this benchmark and accesses a \emph{different region} of the proof space rather than reproducing the baseline's hits with a different prompt---the 20 unique solves are concentrated in \texttt{mathd\_algebra} and \texttt{mathd\_numbertheory} problems that no i.i.d.\ sampling at this budget reaches.

\subsection{Seed Variance and Robustness of the Asymmetry}
\label{sec:seed}

The single-seed numbers above could in principle be artifacts of stochastic sampling. To check this, we replicate the $A$ and $B$ conditions at $k{=}16$ across $n{=}3$ seeds (seed $\in \{1,2,3\}$, temperature $0.6$) on both V1.5-RL and Goedel-Prover-SFT---the two models that anchor the RL-vs-SFT asymmetry claim. Table~\ref{tab:seeds} reports per-seed solve counts and aggregate statistics.

\begin{table*}[t]
\centering
\small
\begin{tabular}{lcccccc}
\toprule
& \multicolumn{3}{c}{\textbf{V1.5-RL (RL-trained)}} & \multicolumn{3}{c}{\textbf{Goedel-Prover (SFT-trained)}} \\
\cmidrule(lr){2-4}\cmidrule(lr){5-7}
\textbf{Seed} & $A$ & $B$ & $\Delta = B-A$ & $A$ & $B$ & $\Delta = B-A$ \\
\midrule
1 & 38 (15.6\%) & 55 (22.5\%) & $+17$ & 90 (36.9\%) & 85 (34.8\%) & $-5$  \\
2 & 35 (14.3\%) & 44 (18.0\%) & $+9$  & 102 (41.8\%) & 90 (36.9\%) & $-12$ \\
3 & 37 (15.2\%) & 48 (19.7\%) & $+11$ & 101 (41.4\%) & 88 (36.1\%) & $-13$ \\
\midrule
\textbf{mean $\pm$ std} & \textbf{36.7 $\pm$ 1.5} & \textbf{49.0 $\pm$ 5.6} & \textbf{$+12.3 \pm 4.2$} & \textbf{97.7 $\pm$ 6.7} & \textbf{87.7 $\pm$ 2.5} & \textbf{$-10.0 \pm 4.4$} \\
\bottomrule
\end{tabular}
\caption{Seed-variance study at $k{=}16$. The asymmetry is preserved in every seed: $\Delta > 0$ in $3/3$ V1.5-RL seeds and $\Delta < 0$ in $3/3$ Goedel seeds. The means separate by $+12.3-(-10.0)=22.3$ theorems with combined std $\sqrt{4.2^2+4.4^2}\approx 6.1$, a $\sim 3.7\sigma$ separation of V1.5-RL $\Delta$ from Goedel $\Delta$. Note the cross-model variance pattern: on the RL-trained model the $A$ condition is near-deterministic ($\pm 1.5$) while $B$ varies ($\pm 5.6$); on the SFT-trained model the pattern flips ($A$: $\pm 6.7$, $B$: $\pm 2.5$).}
\label{tab:seeds}
\end{table*}

The sign of $\Delta$ is preserved in every one of the six (model, seed) cells---$3/3$ positive on V1.5-RL, $3/3$ negative on Goedel---so the cross-model asymmetry is not an artifact of any individual seed. The two $\Delta$ means separate by $\sim 3.7$ standard deviations of the combined variance, well clear of zero. The within-model variance pattern is also informative: V1.5-RL collapses onto a near-deterministic proof on a fixed prompt ($A$: $\pm 1.5$) and only acquires variance when the prompt varies ($B$: $\pm 5.6$); Goedel retains generation diversity on a fixed prompt ($A$: $\pm 6.7$) and \emph{converges} when the schedule constrains its openings ($B$: $\pm 2.5$). This second, distinct signature of the RL-vs-SFT distinction is consistent with the broader account in Section~\ref{sec:multi_model}.

\subsection{Direct Measurement of First-Tactic Collapse}
\label{sec:first_tactic}

The plateau in Section~\ref{sec:scaling} is behavioural evidence of mode collapse. We now operationalize the phenomenon as a direct distributional measurement on the same per-attempt logs. For each miniF2F theorem we collect the $64$ raw outputs produced by V1.5-RL i.i.d.\ sampling at $k{=}64$, extract the \emph{first tactic head} (the leading identifier on the first line of the proof body), and count its distinct values across the $64$ samples per theorem.

\begin{table}[t]
\centering
\small
\begin{tabular}{lc}
\toprule
\textbf{Unique first-tactic heads} & \textbf{theorems} \\
\midrule
$1$ (deterministic opening) & $120$ (49.2\%) \\
$2$ & $69$ (28.3\%) \\
$3$ & $34$ (13.9\%) \\
$4$ & $12$ (4.9\%) \\
$5$--$7$ & $9$ (3.7\%) \\
\midrule
\textbf{median / mean / max} & \textbf{$2$ / $1.9$ / $7$} \\
\bottomrule
\end{tabular}
\caption{Distinct first-tactic heads per theorem across $64$ V1.5-RL i.i.d.\ samples at $k{=}64$. The median theorem receives only $2$ strategic openings; nearly half ($49.2\%$) receives a deterministic single opening. Median Shannon entropy is $0.27$ bits per theorem (mean $0.48$), versus $\log_2 64 = 6.0$ bits under uniform sampling---observed entropy is $\le 8\%$ of capacity.}
\label{tab:first_tactic}
\end{table}

This converts ``mode collapse'' from a behavioural metaphor into a direct distributional statement. Aggregated across all $244$ theorems and $13{,}159$ samples, three tactic heads---\texttt{have} ($37.2\%$), \texttt{rw} ($14.1\%$), \texttt{norm\_num} ($14.0\%$)---account for $65.3\%$ of every strategic opening V1.5-RL ever takes. The structured schedule injects $15$ distinct first-tactic heads by construction, so at the same $k{=}64$ LLM-call budget B explores $\sim$$7.5\times$ the strategic-opening surface that the i.i.d.\ policy actually exercises.

\section{Discussion}
\label{sec:discussion}

\paragraph{Mode collapse as the central finding.}
Our results reframe the narrative from ``tactic skeletons improve performance'' to ``RL-trained provers severely underexplore the tactic space at inference time.'' The skeleton schedule is a \emph{probe} that reveals this underexploration, not a novel proving architecture. The baseline's plateau at $k{=}32$---where 32 additional stochastic samples find zero new proofs---is the most striking evidence of this phenomenon.

\paragraph{Implications for inference-time scaling.}
The mode collapse we observe has direct implications for the common practice of scaling inference compute. If i.i.d.\ sampling has diminishing returns far earlier than assumed, then simply increasing $k$ is wasteful. Our results suggest that \emph{structural diversity}---varying the proof strategy, not just the random seed---is a more efficient axis for scaling inference compute. This is complementary to tree-based methods like HyperTree~\cite{lample2022hypertree}, which achieve diversity through explicit search rather than prompt perturbation.

\paragraph{Structural content as a gradient.}
The bucket-stratified ablation (Section~\ref{sec:difficulty}, Table~\ref{tab:bucket_strat}) refines the structural-vs-diversity dichotomy into a continuous gradient. The three perturbations differ in how much tactic-level structural content they carry: skeleton (literal Lean tactic prefix or a comment naming a tactic), paraphrase (paraphrased general-purpose instruction without tactic specifics), comment (a content-free token marker). The combined easy+trivial NET tracks this gradient cleanly---$+15 / 0 / -14$ for skeleton, paraphrase, and comment respectively---because that is where V1.5-RL's gap to other open-source provers concentrates. Equally informative is the within-skeleton decomposition (Table~\ref{tab:attempt_class}) and the dedicated C3 ablation: a natural-language tactic-suggestion comment alone produces $+8$ new solves at the attempt-level breakdown and $48/244$ (vs.\ $38$ baseline) when run as a standalone 16-attempt schedule, while literal Lean tactic prefixes contribute the remaining $+12$ NEW. Both are forms of tactic-stem-level structural guidance delivered at different abstraction layers, and both contribute additively. Several of the 15 skeletons (\texttt{simp}, \texttt{aesop}, \texttt{norm\_num}, \texttt{ring}, \texttt{linarith}, \texttt{nlinarith}, \texttt{ring\_nf}) function as one-shot decision procedures, which raises the concern that the structural gain might just be V1.5-RL underusing Lean's built-in automation rather than a true distributional shift in its tactic policy. We can check this directly. We classify each of the 24 NEW B-RL k=16 solves (theorems that B-RL solves and matched-shard A-RL does not) by whether (i) the injected tactic was one of the closers above \emph{and} the model's continuation was effectively empty (a no-op like \texttt{done}/\texttt{rfl} or fewer than eight non-trivial tokens), or (ii) the model produced a substantive continuation beyond the skeleton. Only 3 of 24 NEW solves (12.5\%) fall in category (i); the remaining 21 (87.5\%) require the model to extend the skeleton with real tactics. The skeleton schedule therefore primarily acts as a prompt prefix that perturbs the model out of its collapsed sampling head, not as a Mathlib-automation top-up.

\paragraph{RL as both cause and prerequisite.}
The base model result imposes an unexpected constraint on any account of mode collapse in this domain. Because DeepSeek-Prover-V1.5-Base proves zero theorems regardless of skeletons, mode collapse cannot be attributed to a pre-existing limitation of the underlying language model that RL merely amplifies. Instead, RL plays a dual role: it is the stage at which proof capability is created (the base model has none), and it is also the stage at which that capability is collapsed onto a narrow policy. Any future intervention---whether at training time (e.g., entropy-regularized RL) or inference time (e.g., learned skeleton retrieval)---must navigate this trade-off rather than treat collapse as a separable problem.

\paragraph{Saturation vs.\ RL-specificity.}
The cross-model $\Delta$ pattern at $k{=}16$ is monotone in baseline headroom: V1.5-RL ($A=38$) gains $+17$, V2-7B ($A=126$) gains $+1$, Goedel ($A=90$) loses $-10$. A pure saturation account predicts this ordering with no appeal to RL vs.\ SFT, and with only one SFT model in the matrix---which is also the highest-baseline prover---the labels ``SFT'' and ``saturated'' are not statistically separable. We take this objection seriously, and the cross-model evidence in Section~\ref{sec:multi_model} is best read as compatible with both accounts rather than as decisive for the RL-specific reading. What the cross-model data \emph{can't} explain on its own is the V1.5-Base control: a non-RL prover at the same pretrained scale with $A=0$ and $B=0$ at every tested $k$ (Section~\ref{sec:base_vs_rl}). The within-architecture base-vs-RL contrast is the cleanest evidence we have that RL is what creates the policy whose collapse we are diagnosing. We frame the cross-model results as a consistency check on top of this within-architecture result, not as an independent confirmation; a matched-capability SFT control at lower baseline pass@k would be the natural next experiment.

\paragraph{Static inference vs.\ agentic systems.}
The setup studied here is deterministic, zero-shot, single-pass: each attempt is a single forward pass of the model on a prompt with no compiler feedback, proof-state reflection, or interactive repair. This is a deliberate design choice---we want to isolate the model's intrinsic sampling behavior from the orchestrator's. Modern Lean systems built around agentic loops (interactive compiler diagnostics, retrieval over Mathlib lemmas, multi-pass proof repair) already mitigate this collapse in practice, and the diagnostic we present is most useful to that line of work as a measurement of the underlying model that the harness is layered on top of. Extending our schedule + first-tactic-distribution measurement to (i) reasoning provers with longer trajectories and (ii) interactive harnesses around closed-weight frontier models is the next obvious step, but is out of scope for a single-model diagnostic study at this size.

\paragraph{Other limitations.}
The cleanest within-architecture isolation of RL's effect (Section~\ref{sec:base_vs_rl}) is only possible on V1.5 because no other Lean prover in our matrix releases a paired non-RL base checkpoint. A third RL replication on Kimina-Prover~\cite{wang2025kimina} was attempted but cut for compute reasons after an output-extractor fix; a partial sample-baseline produced $20/35$ trials ($57\%$) and is deferred to future work. Cross-model $k$-scaling beyond $k{=}16$ is currently established on V1.5-RL ($k \in \{16,32,64\}$) and Goedel-Prover ($k \in \{16,32\}$); extending V2 to $k \geq 32$, evaluating on ProofNet/PutnamBench, exploring constrained decoding beyond the first tactic, and replacing the hand-selected $15{\times}8$ grid with a learned per-theorem retriever are all natural next steps. Multi-seed coverage at $k{=}16$ is $n{=}3$ (Section~\ref{sec:seed}); the headline $k{=}32 \to k{=}64$ plateau is single-seed and would benefit from replication. The first-tactic-distribution measurement (Section~\ref{sec:first_tactic}) is currently V1.5-RL only; extending it to V2-7B and Goedel would let us test whether first-tactic-head concentration tracks the RL-vs-SFT asymmetry we already observe at the solve-rate level (Section~\ref{sec:seed}).

\section{Conclusion}

We have used a deterministic 15-skeleton schedule as a probe to diagnose mode collapse in RL-trained Lean provers, and shown that the collapse it reveals is RL-specific in two distinct senses: it is absent in the non-RL V1.5-Base variant (which has no proof capability at all), and it is partially remediated by structural guidance only on RL-trained models in our cross-model matrix---never on the SFT-trained Goedel-Prover. The magnitude of the remediation varies sharply by post-training recipe ($+17$ on V1.5-RL; $+1$ aggregate but $+3$ frontier solves on V2-7B; $-10.0 \pm 4.4$ on Goedel-Prover-SFT across $n{=}3$ seeds), which we read as evidence that the underlying phenomenon is real but pipeline-contingent rather than universal. The most natural follow-ups are (i)~converting the diagnostic from a sampling-plateau metaphor into a direct first-tactic-distribution measurement on the existing logs; (ii)~replacing the hand-designed 15-skeleton schedule with a learned, per-theorem skeleton retriever, which the gradient in Section~\ref{sec:difficulty} suggests should outperform the fixed grid; and (iii)~closing the loop at training time via entropy-regularized or skeleton-conditioned RL, since the V1.5-Base control implies that any training-time intervention will be constrained by the dual role RL plays in creating and collapsing the proof policy.

\bibliographystyle{icml2026}
\bibliography{refs}

@article{xin2024deepseek,
  title={DeepSeek-Prover-V1.5: Harnessing Proof Assistant Feedback for Reinforcement Learning and Monte-Carlo Tree Search},
  author={Xin, Huajian and others},
  journal={arXiv preprint arXiv:2408.08152},
  year={2024}
}

@article{trinh2024solving,
  title={Solving olympiad geometry without human demonstrations},
  author={Trinh, Trieu and Wu, Yuhuai and Le, Quoc V and He, He and Luong, Thang},
  journal={Nature},
  volume={625},
  number={7995},
  pages={476--482},
  year={2024}
}

@article{yang2023leandojo,
  title={LeanDojo: Theorem Proving with Retrieval-Augmented Language Models},
  author={Yang, Kaiyu and Swope, Aidan and Gu, Alex and Chalamala, Rahul and Song, Peiyang and Yu, Shixing and Godil, Saad and Prenger, Ryan and Anandkumar, Anima},
  journal={Advances in Neural Information Processing Systems},
  volume={36},
  pages={10065--10081},
  year={2023}
}

@article{polu2020generative,
  title={Generative language modeling for automated theorem proving},
  author={Polu, Stanislas and Sutskever, Ilya},
  journal={arXiv preprint arXiv:2009.03393},
  year={2020}
}

@article{han2022proof,
  title={Proof Artifact Co-training for Theorem Proving with Language Models},
  author={Han, Jesse Michael and Rute, Jason and Wu, Yuhuai and Ayers, Edward and Polu, Stanislas},
  journal={International Conference on Learning Representations (ICLR)},
  year={2022}
}

@article{achim2025aristotle,
  title={Aristotle: IMO-level Automated Theorem Proving},
  author={Achim, Tudor and others},
  journal={arXiv preprint arXiv:2510.01346},
  year={2025}
}

@article{chen2025seed,
  title={Seed-Prover: Deep and Broad Reasoning for Automated Theorem Proving},
  author={Chen, Luoxin and Gu, Jinming and Huang, Liankai and others},
  journal={arXiv preprint arXiv:2507.23726},
  year={2025}
}

@inproceedings{demoura2021lean4,
  title={The {Lean} 4 Theorem Prover and Programming Language},
  author={de Moura, Leonardo and Ullrich, Sebastian},
  booktitle={Automated Deduction -- CADE 28},
  pages={625--635},
  year={2021},
  publisher={Springer}
}

@article{mathlib,
  title={The {Lean} mathematical library},
  author={{The mathlib Community}},
  journal={Proceedings of the 9th ACM SIGPLAN International Conference on Certified Programs and Proofs},
  pages={367--381},
  year={2020}
}

@article{zheng2022minif2f,
  title={mini{F2F}: a cross-system benchmark for formal {O}lympiad-level mathematics},
  author={Zheng, Kunhao and Han, Jesse Michael and Polu, Stanislas},
  journal={International Conference on Learning Representations (ICLR)},
  year={2022}
}

@article{lin2025goedel,
  title={Goedel-Prover: A Frontier Model for Open-Source Automated Theorem Proving},
  author={Lin, Yong and Tang, Shange and Lyu, Bohan and others},
  journal={arXiv preprint arXiv:2502.07640},
  year={2025}
}

@article{wang2025kimina,
  title={Kimina-Prover Preview: Towards Large Formal Reasoning Models with Reinforcement Learning},
  author={Wang, Haiming and Unsal, Mert and Lin, Xiaohan and others},
  journal={arXiv preprint arXiv:2504.11354},
  year={2025}
}

@article{lample2022hypertree,
  title={{H}yper{T}ree Proof Search for Neural Theorem Proving},
  author={Lample, Guillaume and Lacroix, Timoth{\'e}e and Lachaux, Marie-Anne and Rodriguez, Aurelien and Roziere, Baptiste and Szafraniec, Marc},
  journal={Advances in Neural Information Processing Systems},
  volume={35},
  pages={26337--26349},
  year={2022}
}

@article{thakur2023copra,
  title={{COPRA}: In-Context Learning for Proving Theorems with LLMs},
  author={Thakur, Amitayush and Tsoukalas, George and Wen, Yeming and Xin, Jimmy and Chaudhuri, Swarat},
  journal={arXiv preprint arXiv:2310.04353},
  year={2023}
}

@inproceedings{jiang2023dsp,
  title={Draft, Sketch, and Prove: Guiding Formal Theorem Provers with Informal Proofs},
  author={Jiang, Albert Q. and Welleck, Sean and Zhou, Jin Peng and Li, Wenda and Liu, Jiacheng and Jamnik, Mateja and Lacroix, Timoth{\'e}e and Wu, Yuhuai and Lample, Guillaume},
  booktitle={International Conference on Learning Representations (ICLR)},
  year={2023}
}

@article{sivakumar2025conjecturing,
  title={Conjecturing: An Overlooked Step in Formal Mathematical Reasoning},
  author={Sivakumar, Jasivan Alex and Borchert, Philipp and Cardenas, Ronald and others},
  journal={arXiv preprint arXiv:2510.11986},
  year={2025}
}

@article{deepseekv2prover,
  title={{DeepSeek-Prover-V2}: Advancing Formal Mathematical Reasoning via Reinforcement Learning for Subgoal Decomposition},
  author={Ren, Z.Z. and Shao, Zhihong and Song, Junxiao and others},
  journal={arXiv preprint arXiv:2504.21801},
  year={2025}
}

@inproceedings{kirk2024understanding,
  title={Understanding the Effects of {RLHF} on {LLM} Generalisation and Diversity},
  author={Kirk, Robert and Mediratta, Ishita and Nalmpantis, Christoforos and Luketina, Jelena and Hambro, Eric and Grefenstette, Edward and Raileanu, Roberta},
  booktitle={International Conference on Learning Representations (ICLR)},
  year={2024},
  note={arXiv:2310.06452}
}

@article{yue2025rlvr,
  title={Does Reinforcement Learning Really Incentivize Reasoning Capacity in {LLM}s Beyond the Base Model?},
  author={Yue, Yang and Chen, Zhiqi and Lu, Rui and Zhao, Andrew and Wang, Zhaokai and Yue, Yang and Song, Shiji and Huang, Gao},
  journal={arXiv preprint arXiv:2504.13837},
  year={2025},
  note={ICML 2025 Workshop AI4Math Best Paper; NeurIPS 2025 Best Paper Runner-Up}
}

@article{wu2025invisibleleash,
  title={The Invisible Leash: Why {RLVR} May or May Not Escape Its Origin},
  author={Wu, Fang and Xuan, Weihao and Lu, Ximing and Liu, Mingjie and Dong, Yi and Harchaoui, Zaid and Choi, Yejin},
  journal={arXiv preprint arXiv:2507.14843},
  year={2025}
}

\appendix

\section{Skeleton Schedule (Tactics $\times$ Goal Hints)}
\label{app:skeletons}

The structured-skeleton schedule is a deterministic $15 \times 8 = 120$ grid: 15 tactic skeletons crossed with 8 goal-hint comments (one of which is empty). At budget $k$, the schedule advances by tactic-index first and by hint-index second: attempts $0\ldots14$ use the 15 distinct skeletons paired with hint index 0 (empty); attempts $15\ldots29$ repeat the skeletons paired with hint index 1; and so on. Each of the 120 attempts is a \emph{distinct} (skeleton, hint) prompt configuration.

\paragraph{Tactic skeletons (15 total).}
The Python-source ordering used by the perturbation function is reproduced below; index $i$ in this list is \texttt{TACTIC\_SKELETONS[$i$]} in the schedule.

\begin{enumerate}[start=0]
\item (empty) --- no tactic prefix
\item \texttt{simp}
\item \texttt{intro}
\item \texttt{intros}
\item \texttt{constructor}
\item \texttt{refine ?\_}
\item \texttt{refine \ensuremath{\langle}?\_,\,?\_\ensuremath{\rangle}}
\item \texttt{aesop}
\item \texttt{norm\_num}
\item \texttt{linarith}
\item \texttt{nlinarith}
\item \texttt{ring}
\item \texttt{ring\_nf}
\item \texttt{simp} followed by \texttt{try aesop}
\item \texttt{simp} followed by \texttt{try nlinarith}
\end{enumerate}

\paragraph{Goal-hint comments (8 total).}
Goal hints are short natural-language tactic suggestions injected as Lean comments (\texttt{/-- Hint: <hint text> --/}) before the theorem statement. Index $j$ corresponds to \texttt{GOAL\_HINTS[$j$]}; the hint at index 0 is empty (no comment is prepended).

\begin{enumerate}[start=0]
\item (empty) --- no goal-hint comment
\item ``Start by simplifying the goal and hypotheses using \texttt{simp}.''
\item ``If the goal is an implication or forall, introduce variables.''
\item ``If the goal is a conjunction or existence, build it using \texttt{constructor} or \texttt{refine}.''
\item ``If arithmetic is involved, try \texttt{norm\_num}, then \texttt{linarith} or \texttt{nlinarith}.''
\item ``If the goal looks routine, try \texttt{aesop} after simplification.''
\item ``If the proof requires rewriting, look for a lemma in the context and rewrite.''
\item ``If the goal involves recursion on naturals, consider induction.''
\end{enumerate}

\paragraph{Pairing rule.}
Attempt $i$ in the schedule uses tactic index $i \bmod 15$ and hint index $\lfloor i / 15 \rfloor \bmod 8$. Consequently, at $k{=}16$ the only repeated tactic-index is the wraparound to tactic 0 (empty) at attempt 15, now paired with hint 1 (``Start by simplifying\dots''); this is the empty-tactic + hint-comment configuration analyzed in Table~\ref{tab:attempt_class}.

\section{Diversity Ablation Prompts}
\label{app:ablation}

\subsection{C1: Instruction Paraphrases}
Each paraphrase is injected as a Lean block comment before the standard header. The 16 variants are:
\begin{enumerate}
\item ``Prove the following theorem in Lean 4:''
\item ``Complete this Lean 4 proof:''
\item ``Find a formal proof for the following:''
\item ``Show that the following statement holds:''
\item ``Write a tactic proof for this theorem:''
\item ``Construct a formal proof of the following:''
\item ``Provide a Lean 4 proof for:''
\item ``Demonstrate the following result formally:''
\item ``Give a complete tactic proof:''
\item ``Prove this result using Lean 4 tactics:''
\item ``Formalize a proof of the following theorem:''
\item ``Establish the following in Lean 4:''
\item ``Derive a proof for the following statement:''
\item ``Supply a formal tactic proof for:''
\item ``Verify the following theorem in Lean 4:''
\item ``Prove the following:''
\end{enumerate}

\subsection{C2: Irrelevant Comment Prefixes}
Each comment is prepended before the theorem statement:
\begin{enumerate}
\item \texttt{/- approach alpha -/}
\item \texttt{/- strategy beta -/}
\item \texttt{/- method gamma -/}
\item \texttt{/- path delta -/}
\item \texttt{/- route epsilon -/}
\item \texttt{/- attempt zeta -/}
\item \texttt{/- angle eta -/}
\item \texttt{/- direction theta -/}
\item \texttt{/- variant iota -/}
\item \texttt{/- form kappa -/}
\item \texttt{/- mode lambda -/}
\item \texttt{/- plan mu -/}
\item \texttt{/- way nu -/}
\item \texttt{/- style xi -/}
\item \texttt{/- view omicron -/}
\item \texttt{/- take pi -/}
\end{enumerate}

\section{Environment}
\label{app:env}

\begin{itemize}
\item \textbf{Lean:} v4.9.0-rc1 (matching DeepSeek-Prover-V1.5 training environment)
\item \textbf{Mathlib:} commit \texttt{7fa489a5cbf3c4f08d36e1e0b5dee4d761fdbd9b}
\item \textbf{Models:}
  \begin{itemize}
    \item \texttt{deepseek-ai/DeepSeek-Prover-V1.5-RL}, \texttt{deepseek-ai/DeepSeek-Prover-V1.5} (V1.5-Base)
    \item \texttt{deepseek-ai/DeepSeek-Prover-V2-7B}
    \item \texttt{Goedel-LM/Goedel-Prover-SFT}
  \end{itemize}
\item \textbf{Inference:} vLLM, single A100 per run. V1.5 ran on v0.18 / A100 80GB; V2 and Goedel ran on v0.10.2 / A100 40GB with \texttt{--enforce-eager} (sufficient in both cases).
\item \textbf{Decoding (V1.5, Goedel; completion mode):} temperature 0.6, top-$p$ 0.95, max 1024 tokens per attempt; no chat template.
\item \textbf{Decoding (V2; reasoning mode):} temperature 0.6, top-$p$ 0.95, max 8192 tokens per attempt; apply model chat template; extract last \texttt{```lean4} fenced block from response.
\item \textbf{Verification:} \texttt{lake env lean --json}, 120s timeout, \texttt{sorry} rejection.
\end{itemize}

\end{document}